\documentclass[twoside,11pt]{article}

\usepackage[abbrvbib]{jmlr2e}
\usepackage{hyperref}

\usepackage{graphicx,wrapfig,lipsum}

\usepackage{listings}
\usepackage{color}

\definecolor{dkgreen}{rgb}{0,0.6,0}
\definecolor{gray}{rgb}{0.5,0.5,0.5}
\definecolor{mauve}{rgb}{0.58,0,0.82}

\ShortHeadings{Darts: User-Friendly Modern Machine Learning for Time Series}{}
\firstpageno{1}

\usepackage{lastpage}
\jmlrheading{23}{2022}{1-\pageref{LastPage}}{10/21; Revised
2/22}{3/22}{21-1177}{
Julien Herzen,
Francesco Lässig,
Samuele Giuliano Piazzetta,
Thomas Neuer,
Léo Tafti,
Guillaume Raille,
Tomas Van Pottelbergh,
Marek Pasieka,
Andrzej Skrodzki,
Nicolas Huguenin,
Maxime Dumonal,
Jan Kościsz,
Dennis Bader,
Frédérick Gusset,
Mounir Benheddi,
Camila Williamson,
Michal Kosinski,
Matej Petrik,
Gaël Grosch
}
\ShortHeadings{Darts: User-Friendly Modern Machine Learning for Time Series}{
Herzen, Lässig, Piazzetta, Neuer et al.
}

\begin{document}

\title{Darts: User-Friendly Modern Machine Learning for Time Series}

\author{\name Julien Herzen\dag \email julien.herzen@unit8.com\\
        \name Francesco Lässig\dag \email francesco.laessig@unit8.com\\
        \name Samuele Giuliano Piazzetta \email samuele.piazzetta@gmail.com\\
        \name Thomas Neuer\dag \email thomas.neuer@unit8.com\\
        \name Léo Tafti \email leotafti@gmail.com\\
        \name Guillaume Raille\dag \email guillaume.raille@unit8.com\\
        \name Tomas Van Pottelbergh\dag \email tomas.vanpottelbergh@unit8.com\\
        \name Marek Pasieka\dag \email marek.pasieka@unit8.com\\
        \name Andrzej Skrodzki \email andrzej.skrodzki@gmail.com\\
        \name Nicolas Huguenin\dag \email nicolas.huguenin@unit8.com\\
        \name Maxime Dumonal\dag \email maxime.dumonal@unit8.com\\
        \name Jan Kościsz \email koscis.j@gmail.com\\
        \name Dennis Bader\dag \email dennis.bader@unit8.com\\
        \name Frédérick Gusset \email gusset.frederick@gmail.com\\
        \name Mounir Benheddi \email mounir.benheddi@gmail.com\\
        \name Camila Williamson \email camilaagw@gmail.com\\
        \name Michal Kosinski \email kosinski.michal@gmail.com\\
        \name Matej Petrik \email matej.petrik@unit8.com\\
        \name Gaël Grosch\dag \email gael.grosch@unit8.com\\
        \emph{\dag Unit8 SA, Switzerland}}

\editor{Alexandre Gramfort}

\maketitle

\begin{abstract}%   <- trailing '%' for backward compatibility of .sty file
We present Darts\footnote{\href{https://github.com/unit8co/darts}{https://github.com/unit8co/darts}}, 
a Python machine learning library for time series, with a focus on forecasting.
Darts offers a variety of models, from classics such as ARIMA to state-of-the-art deep neural networks.
The emphasis of the library is on offering modern machine learning functionalities, such as 
supporting multidimensional series, fitting models on multiple series, training on large 
datasets, incorporating external data, ensembling models, and providing a rich support for probabilistic forecasting.
At the same time, great care goes into the API design to make it user-friendly and easy to use.
For instance, all models can be used using \texttt{fit()}/\texttt{predict()}, similar to scikit-learn \citep{scikit-learn11}.
\end{abstract}


\section{Introduction}
Time series forecasting has numerous industrial and scientific applications in logistics, predictive maintenance, energy, 
manufacturing, agriculture, healthcare, sales, climate science, and many other domains.
While classical methods such as ARIMA and Exponential Smoothing often give good results~\citep{hyndman18}, machine learning is becoming a more attractive option to improve the models' representation power
and scale to larger datasets and higher dimensionalities. In fact, it has recently been shown that 
pure ML-based approaches based on generic deep learning architectures
can beat classical methods on a variety of tasks \citep{nbeats19, deepar17}.

Perhaps more important than sheer accuracy, the arrival of modern machine learning opens the opportunity
to re-think the forecasting practices and software.
For example, classical methods typically require training one model per time series, whereas ML models usually 
work best when trained on datasets containing large numbers of time series.
This and other paradigm changes -- such as better support for high-dimensional data, iterative training based on
stochastic gradient descent, or the possibility to tailor loss functions for specific needs -- require new tools and appropriate APIs.
In particular, user-friendly and powerful APIs are important to make ML approaches as easy to use as classical techniques, which is necessary for larger-scale adoption by practitioners.

Several strong time series forecasting toolkits exist; however, they focus on classical methods or do not 
support deep learning and training models on multiple series \citep{rforecasting08, kats21, sktime19, greykite21, merlion21}, 
or have lower-level APIs \citep{gluonts20, ptforecasting20}. Darts proposes a new relatively high-level 
API unifying classical and ML-based forecasting models.

\section{Design Principles and Main Features of Darts}
\subsection{Time Series Representation}
Darts has its own \texttt{TimeSeries} data container type, which represents one time series.
\texttt{TimeSeries} are immutable and provide guarantees that the data represents a 
well-formed time series with correct shape, type, and sorted time index.
\texttt{TimeSeries} can be indexed either with Pandas \texttt{DatetimeIndex} or \texttt{RangeIndex}~\citep{pandas10}. 
The \texttt{TimeSeries} are wrapping around three-dimensional xarray \texttt{DataArray} \citep{xarray17}.
The dimensions are (\emph{time, component, sample}), where \emph{component} represents the dimensions of multivariate
series and \emph{sample} represents samples of stochastic time series.
The \texttt{TimeSeries} class provides several methods to convert to/from other common types, 
such as Pandas Dataframes or NumPy arrays~\citep{numpy20}.
It can also perform convenient operations, such as math operations, indexing, splitting, time-differencing,
interpolating, mapping functions, embedding timestamps, plotting, or computing marginal quantiles.
For immutability, \texttt{TimeSeries} carry their own copy of the data and heavily rely on NumPy views 
for efficiently accessing the data without copying (e.g., when training models).

The main advantage of using a dedicated type offering such guarantees is that all models in Darts
can consume and produce \texttt{TimeSeries}, which in turn helps to offer a consistent API.
For instance, it is easy to have models consuming the outputs of other models.

\subsection{Unified High-Level Forecasting API}
All models in Darts support the same basic \texttt{fit(series: TimeSeries)} and \texttt{predict(n: int) -> TimeSeries} 
interface to be trained on a single series \texttt{series} and forecast \texttt{n} time steps after the end of the series.
In addition, most models also provide richer functionalities; for instance the ability to be trained on a \texttt{Sequence}
of \texttt{TimeSeries} (using calls like \texttt{fit([series1, series2, \ldots])}). Models can have different
internal mechanics (e.g., sequence-to-sequence, fixed lengths, recurrent, auto-regressive), and this 
unified API makes it possible to seamlessly compare, backtest, and ensemble diverse models without having to know
their inner workings. 

Some of the models implemented in Darts at the time of writing are: (V)ARIMA, 
Exponential Smoothing, AutoARIMA~\citep{pmdarima17}, Theta \citep{theta00}, Prophet \citep{prophet18}, FFT-based forecasting, 
RNN models similar to DeepAR \citep{deepar17}, N-BEATS \citep{nbeats19}, TCN \citep{tcn18}, TFT \citep{tft21} and
generic regression models that can wrap around any external tabular regression model (such as scikit-learn models).
The list is constantly expanding and we welcome external and reference implementations of new models.

\subsection{Training Models on Collections of Time Series}
An important part of Darts is the support for training one model on a potentially large number of
separate time series \citep{meta-learning21}.
The \texttt{darts.utils.data} module contains various classes implementing
different ways of slicing series (and potential covariates) into training samples.
Darts selects a model-specific default slicing logic, but it can also be user-defined in a custom way if needed. 
All neural networks are implemented using PyTorch \citep{pytorch19} and support training and inference on GPUs . 
It is possible to consume large datasets that do not hold in
memory by relying on custom \texttt{Sequence} implementations to load the data in a lazy fashion.

\subsection{Support for Past and Future Covariates}
Several models support \emph{covariate} series as a way to specify external data potentially helpful for forecasting the target series. 
Darts differentiates \emph{future covariates}, which are known into the future
(such as weather forecasts) from \emph{past covariates}, which are known only into the past.
The models accept \texttt{past\_covariates} and/or \texttt{future\_covariates} arguments,
which make it clear whether future values are required at inference time and reduces the risks
to make mistakes.
Covariate series need not be aligned with the targets, as the alignment is done by the slicing logic
based on the respective time axes.

\subsection{Probabilistic Forecasting}
Some models (and all deep learning models) in Darts support probabilistic forecasting.
The joint distributions over components and time are represented by storing Monte Carlo samples
in the \texttt{TimeSeries} objects directly. This representation is very flexible as it does not
rely on any parametric form and can capture arbitrary joint distributions. The computational cost
of sampling is usually negligible, as samples can be efficiently computed in a vectorized way using batching.
Probabilistic deep learning models can fit arbitrary likelihood forms, as long as the negative log-likelihood is differentiable. 
At the time of writing, Darts provides 17 distributions out-of-the-box
(both continuous and discrete, univariate and multivariate). Finally, it offers 
a way to specify time-independent priors on the distributions' parameters, as a
way to specify prior beliefs about the output distributions.

\subsection{Other Features}
Darts contains many additional features, such as transformers and pipelines for data pre-processing,
backtesting (all models offer a \texttt{backtest()} method), hyperparameter search,
extensive metrics, a dynamic time warping module~\citep{dtw94}, and ensemble models 
(with the possibility to use a regression model to learn the ensemble itself).
Darts also contains \emph{filtering} models such as Kalman filters and
Gaussian Processes, which offer probabilistic modelling of time series.
Finally, the \texttt{darts.datasets} module contains a variety of publicly available time series
which can be conveniently loaded as \texttt{TimeSeries}.

\section{Usage Example}
The code below shows how to fit a single TCN model \citep{tcn18} with default hyper-parameters
on two different (and completely distinct) series,
and forecast one of them. The network outputs the parameters of a Laplace distribution.
The code contains a complete predictive pipeline, from loading and preprocessing the data, 
to plotting the forecast with arbitrary quantiles (shown on the right).

%\vspace{-0.8cm}

\begin{wrapfigure}{r}{0.001mm}
\vspace{2.5cm}  % 3cm for camera ready, 2.5cm otherwise
\hspace*{-6.7cm}
\includegraphics[width=6.65cm]{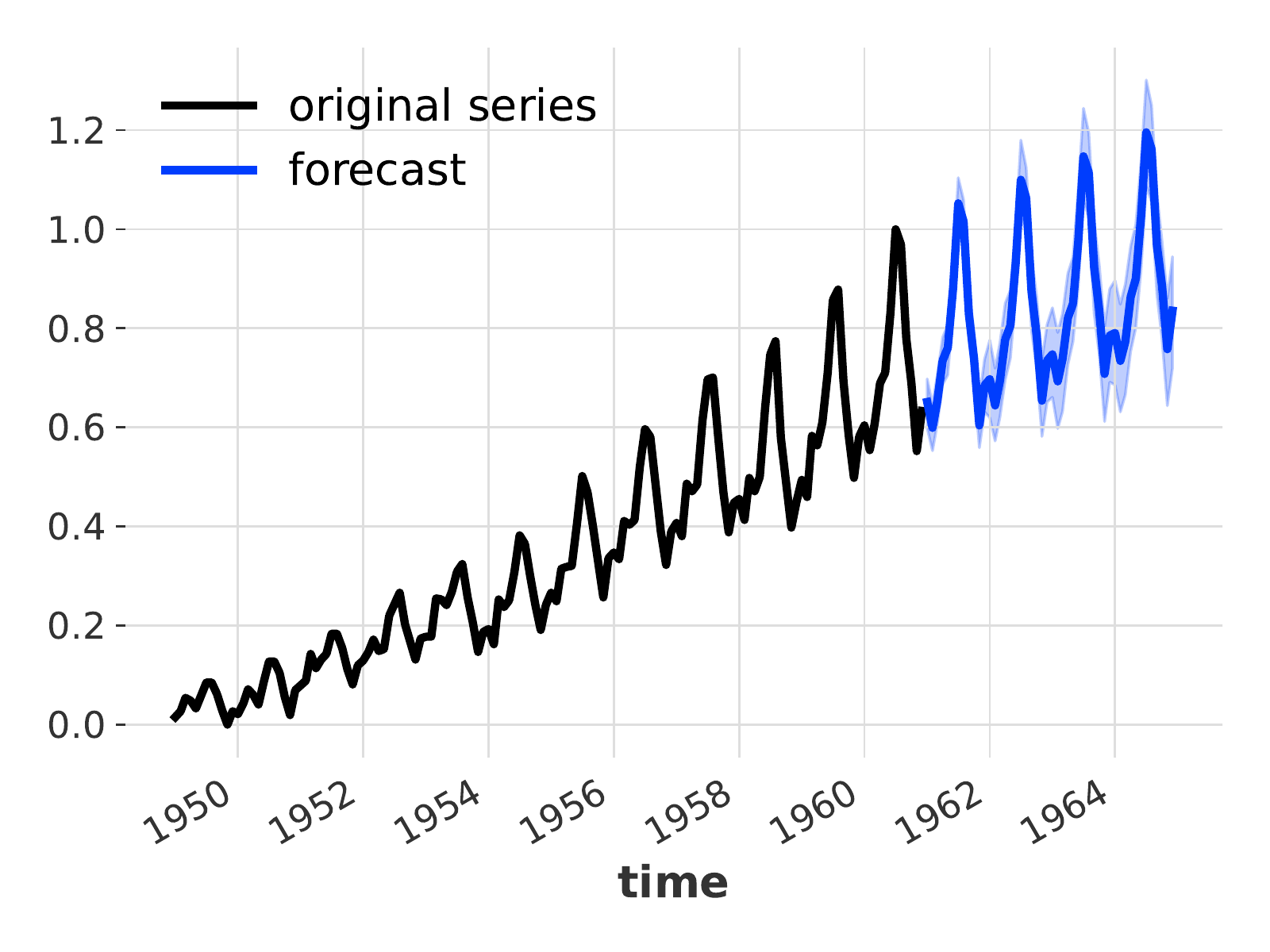}
\end{wrapfigure} 

\lstset{linewidth=\textwidth}
\begin{lstlisting}
from darts.datasets import AirPassengersDataset, MonthlyMilkDataset
from darts.dataprocessing.transformers import Scaler
from darts.models import TCNModel
from darts.utils.likelihood_models import LaplaceLikelihood as LL

air = AirPassengersDataset().load()
milk = MonthlyMilkDataset().load()

scaler_air, scaler_milk = Scaler(), Scaler()
air_s = scaler_air.fit_transform(air)
milk_s = scaler_milk.fit_transform(milk)

model = TCNModel(input_chunk_length=24, 
                   output_chunk_length=12,
                   likelihood=LL())
model.fit([air_s, milk_s], epochs=100)

pred = model.predict(n=48, series=air_s, 
                       num_samples=500)

air_s.plot(label='original series')
pred.plot(low_quantile=.1, high_quantile=.9, label='forecast')
\end{lstlisting}

\section{Conclusions}
Darts is an attempt at democratizing modern machine learning forecasting approaches, and unify them (along with classical
approaches) under a common user-friendly API.
The library is still under active development, and some of the future work includes supporting static covariates and providing a collection of models pre-trained on large datasets, similar to what exists in the computer vision and NLP domains.

%\vskip 0.2in
\bibliography{bibliography.bib}

\end{document}